\documentclass[letterpaper,10pt,conference]{ieeeconf}

\IEEEoverridecommandlockouts
\let\labelindent\relax

\usepackage{url}

\usepackage{color}
\usepackage{amsmath}
\usepackage{accents}
\usepackage{xspace}
\usepackage{enumitem}
\usepackage{algorithm}
\usepackage{algpseudocode}
\usepackage{amsfonts}
\usepackage{cite}
\usepackage{graphicx}
\usepackage{subcaption}
\usepackage[colorlinks]{hyperref}
\usepackage{nameref}

\usepackage{booktabs}
\usepackage{multirow}
\usepackage{dcolumn}
\newcolumntype{d}[1]{D{.}{.}{#1}}
\usepackage[normalem]{ulem}
\usepackage{comment}
\usepackage{amsmath}

\DeclareMathOperator*{\argmin}{arg\,min}

\usepackage{xcolor}

\definecolor{darkgreen}{RGB}{100, 100, 0}

\newcommand{\VS}[1]{#1}

\newtheorem{proof}{Proof}

\usepackage{pifont}

\graphicspath{{figs/}}

\begin{document}

\title{\LARGE \bf MAP-NBV: Multi-agent Prediction-guided Next-Best-View Planning\\ for Active 3D Object Reconstruction}

\author{Harnaik Dhami* \and Vishnu Dutt Sharma* \and Pratap Tokekar
\thanks{*Equal contribution. Names are listed alphabetically.}
\thanks{Authors are with the Department of Computer Science, University of Maryland, U.S.A. \texttt{\small \{dhami, vishnuds, tokekar\}@umd.edu}.}\thanks{This work is supported by the ONR (grant number N00014-18-1-2829).}}

\maketitle
\IEEEpeerreviewmaketitle

\begin{abstract}
    Next-Best View (NBV) planning is a long-standing problem of determining where to obtain the next best view of an object from, by a robot that is viewing the object. There are a number of methods for choosing NBV based on the observed part of the object. In this paper, we investigate how predicting the unobserved part helps with the efficiency of reconstructing the object. We present, Multi-Agent Prediction-Guided NBV (\textit{MAP-NBV}), a decentralized coordination algorithm for active 3D reconstruction with multi-agent systems. Prediction-based approaches have shown great improvement in active perception tasks by learning the cues about structures in the environment from data. However, these methods primarily focus on single-agent systems. We design a decentralized next-best-view approach that utilizes geometric measures over the predictions and jointly optimizes the information gain and control effort for efficient collaborative 3D reconstruction of the object. Our method achieves 19\% improvement over the non-predictive multi-agent approach in simulations using AirSim and ShapeNet. We make our code publicly available through our project website: \url{http://raaslab.org/projects/MAPNBV/}.
\end{abstract}

\section{Introduction}

Visual surveying and inspection with robots has been studied for a long time for a wide range of applications such as inspection of civil infrastructure~\cite{shanthakumar2018view,dhamiGATSBI} and large vehicles~\cite{kim_2009_IROS,ropek_2021}, precision agriculture~\cite{dhami2020crop}, and digital mapping for real estate~\cite{46965,ramakrishnan2021hm3d}. Using robots in these applications is highly advantageous as they can access hard-to-reach areas with greater ease and safety compared to humans. 
 
This work focuses on one such long-studied problem of 3D object reconstruction~\cite{bajcsy2018revisiting}, where the objective is to digitally reconstruct the object of interest by combining observations from multiple vantage points. While it could be easier to achieve this in an indoor environment by carefully placing sensors around the object, the same can not be achieved in the outdoors and open areas. For the latter, the sensor(s), must be moved around the object to capture information from different viewpoints. This can be realized with sensors such as cameras and LiDARs mounted on Unmanned Aerial Vehicles (UAVs). 

A typical goal is to be selective in choosing observation locations for the UAV to achieve an accurate 3D reconstruction as fast as possible. The trade-off between reconstruction accuracy and task duration in unknown environments is commonly addressed through Next-Best-View (NBV) planning~\cite{connolly1985determination}, wherein a robot determines the optimal location for the next observation to maximize information gain. 

Employing a team of UAVs for this problem is an intuitive solution as multiple UAVs can simultaneously cover multiple viewpoints. NBV planning can be used to move the UAVs, gather new observations, and fill the gaps in the object representation. An efficient reconstruction, however, requires not only a correct estimation of the missing information but also coordination among the UAVs to minimize redundancies in the observations.

\begin{figure}[ht!]
\centering
\begin{subfigure}{.47\columnwidth}
    \includegraphics[width=1.0\linewidth]{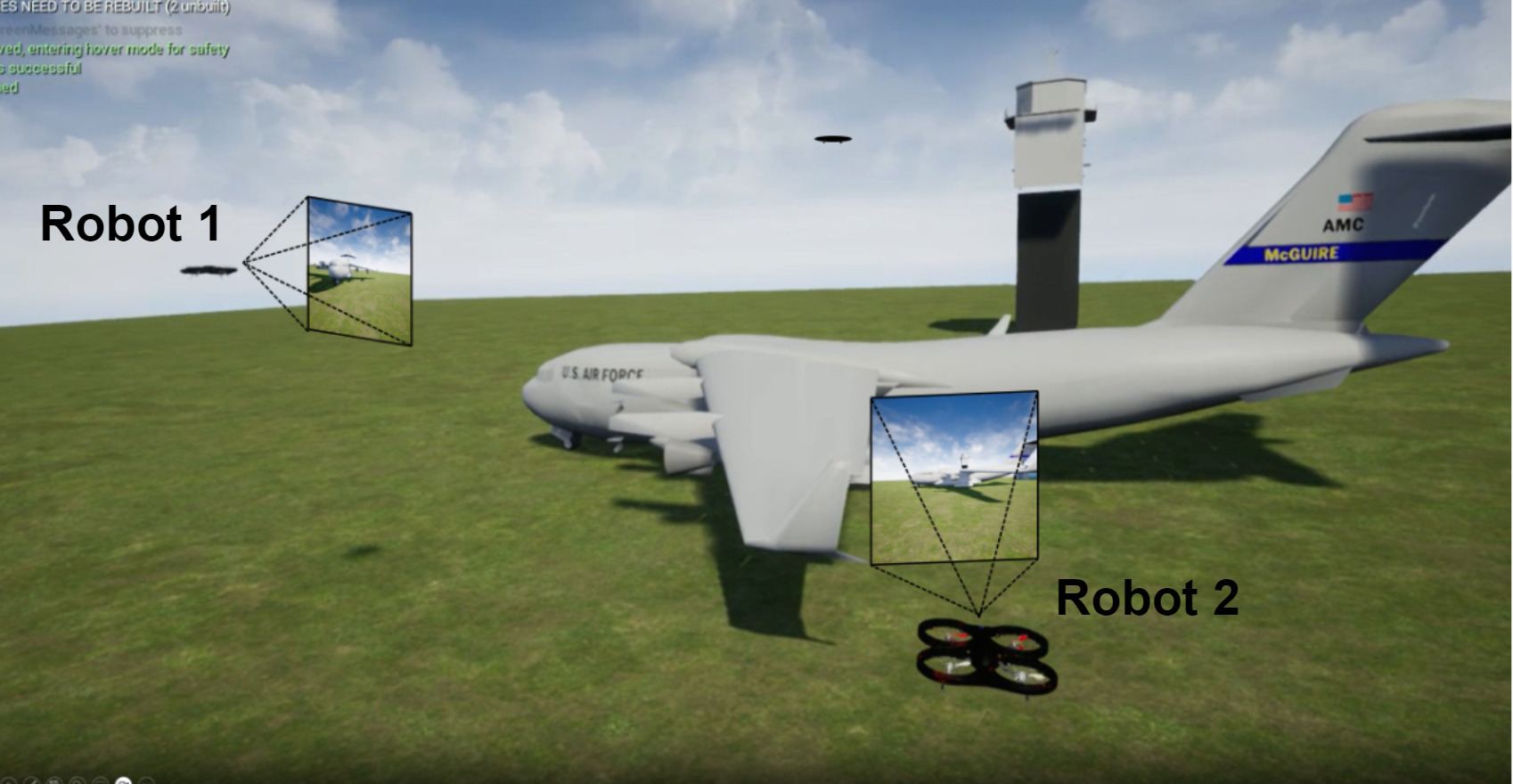}
        \caption{C-17 and the robots in AirSim simulation}
        \label{fig:airsimScreengrab}
\end{subfigure}%
\hfill
\begin{subfigure}{.47\columnwidth}
        \includegraphics[width=0.7\linewidth]{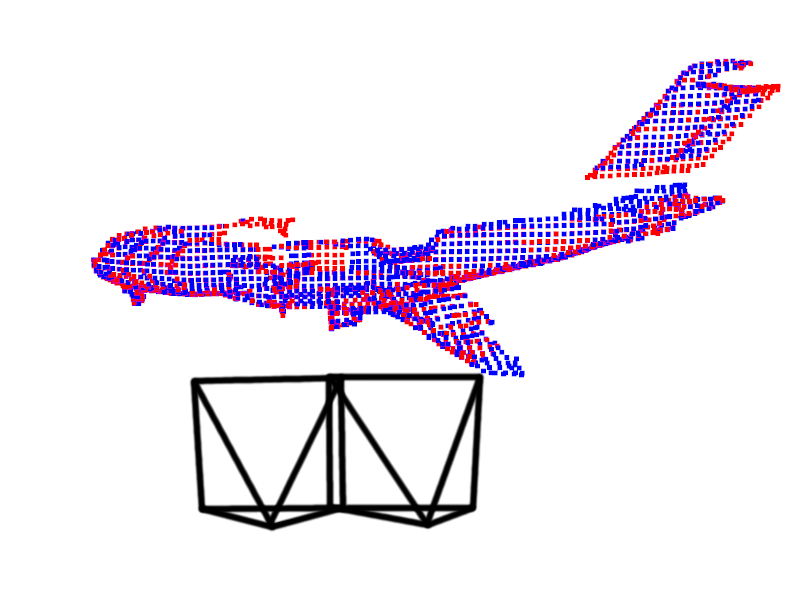}
        \caption{Initial observations by \textcolor{red}{robot 1} and \textcolor{blue}{robot 2}}
        \label{fig:c17Observed}
\end{subfigure}
\medskip
\begin{subfigure}{.47\columnwidth}
        \includegraphics[width=1.0\linewidth]{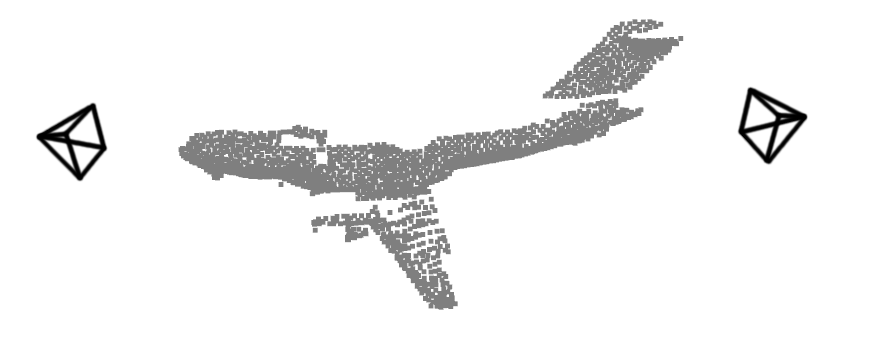}
        \caption{Poses selected by frontiers baseline based on \textcolor{gray}{observations}}
        \label{fig:c17BaselineNBV}
\end{subfigure}%
\hfill
\begin{subfigure}{.47\columnwidth}
        \includegraphics[width=0.97\linewidth]{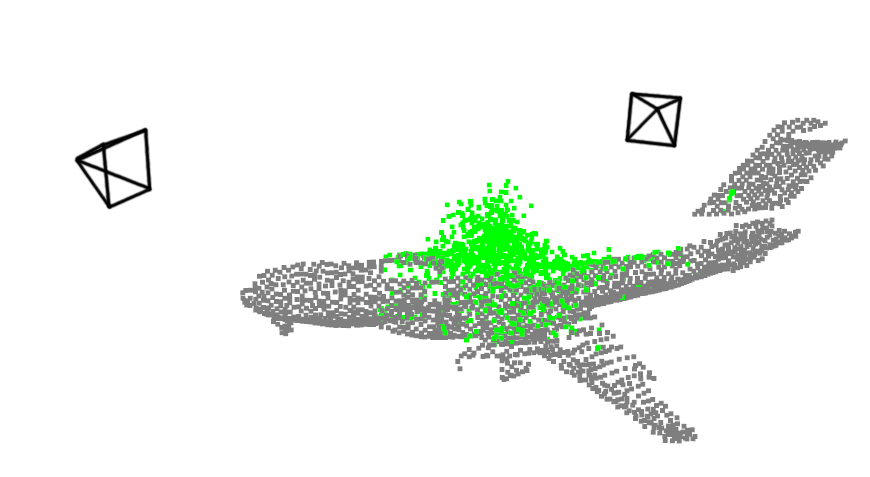}
        \caption{Poses selected by \textit{MAP-NBV} based on \textcolor{gray}{observations}}
        \label{fig:c17MAPNBV}
\end{subfigure}

\medskip
\begin{subfigure}{.47\columnwidth}
        \includegraphics[width=1.0\linewidth]{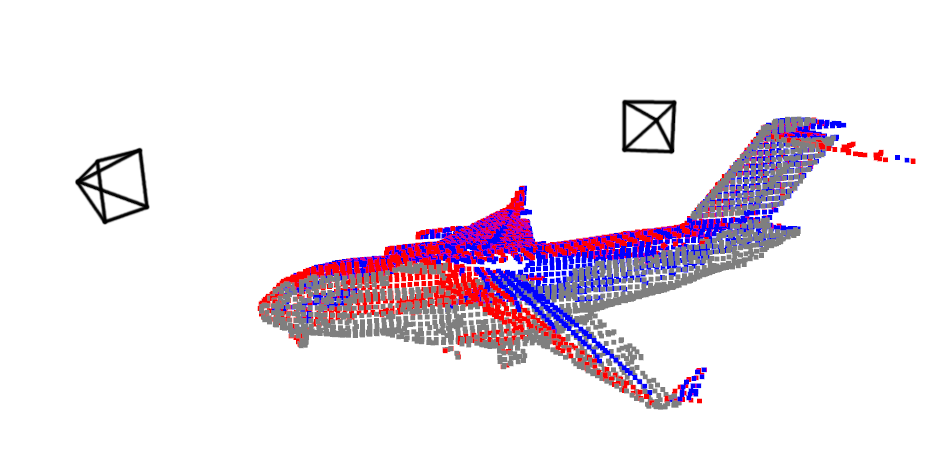}
        \caption{Observations after the first \textit{MAP-NBV} iteration}
        \label{fig:c17BaselineNBV}
\end{subfigure}%
\hfill
\begin{subfigure}{.47\columnwidth}
        \includegraphics[width=0.97\linewidth]{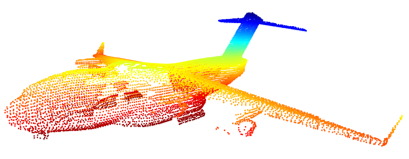}
        \caption{Full point cloud observed by \textit{MAP-NBV} after termination}
        \label{fig:c17MAPNBV}
\end{subfigure}

\caption{MAP-NBV uses \textcolor{green}{predictions} to select better NBVs for a team of robots compared to the non-predictive baseline approach.}
\label{fig:c17predicted}
\label{fig:pointsObserved}
\vspace{-3mm}
\end{figure}

As shown in our prior work, Pred-NBV~\cite{dhami2023prednbv}, estimating the unseen parts of the objects with point cloud completion networks can improve NBV planning, and hence the reconstruction efficiency, for a single UAV. These findings lead us to ask: \textbf{Can prediction improve the efficiency of multi-agent object reconstruction, given that multiple robots can themselves provide good coverage of the object?} 
Assuming predictions can augment the perception of multi-agent systems as well, a naive extension of methods designed for a single agent may result in significant overlaps in the observations by the team, necessitating coordination among all the robots. Prior works have shown that for target coverage problems (which object reconstruction is) \textit{explicit coordination} plays an important role in developing an efficient solution~\cite{corah2019communication}. However, this prior work only focused on scenarios where the coordination used past observation and not predictions. This begs the question: \textbf{How does coordination, in perception and planning, affect multi-agent object reconstruction when each robot has access to predictions?}

\VS{To answer these questions, we make the following contributions in this work}:
\begin{enumerate}
    \item We propose a decentralized, multi-agent, prediction-based NBV planning approach, named\textit{ MAP-NBV}, for active 3D reconstruction of various objects with a novel objective combining visual information gain and control effort.\looseness=-1
    
    \textit{MAP-NBV} uses partial point clouds and predicts what the rest of the point cloud would be (Figure~\ref{fig:c17predicted}) and exploits the submodular nature of the objective to coordinate in a decentralized fashion.
    \item We show that predictions effectively improve the performance by \textbf{19.41\%} over non-predictive baselines that use frontier-based NBV planning~\cite{aleotti2014global} in AirSim~\cite{shah2018airsim} simulations.\looseness=-1
    \item We also show that \textit{MAP-NBV} results in at least \textbf{17.12\%} better reconstruction than non-cooperative prediction-guided method with experiments using the ShapeNet~\cite{shapenet2015} dataset and performs comparable to a centralized approach.\looseness=-1

\end{enumerate}
We share the qualitative results and release the project code from our method on our project website.\footnote{\url{http://raaslab.org/projects/MAPNBV/}}

\begin{figure*}[ht!]{}
    \vspace{1mm}
    \centering
    \includegraphics[width = 1.98\columnwidth]{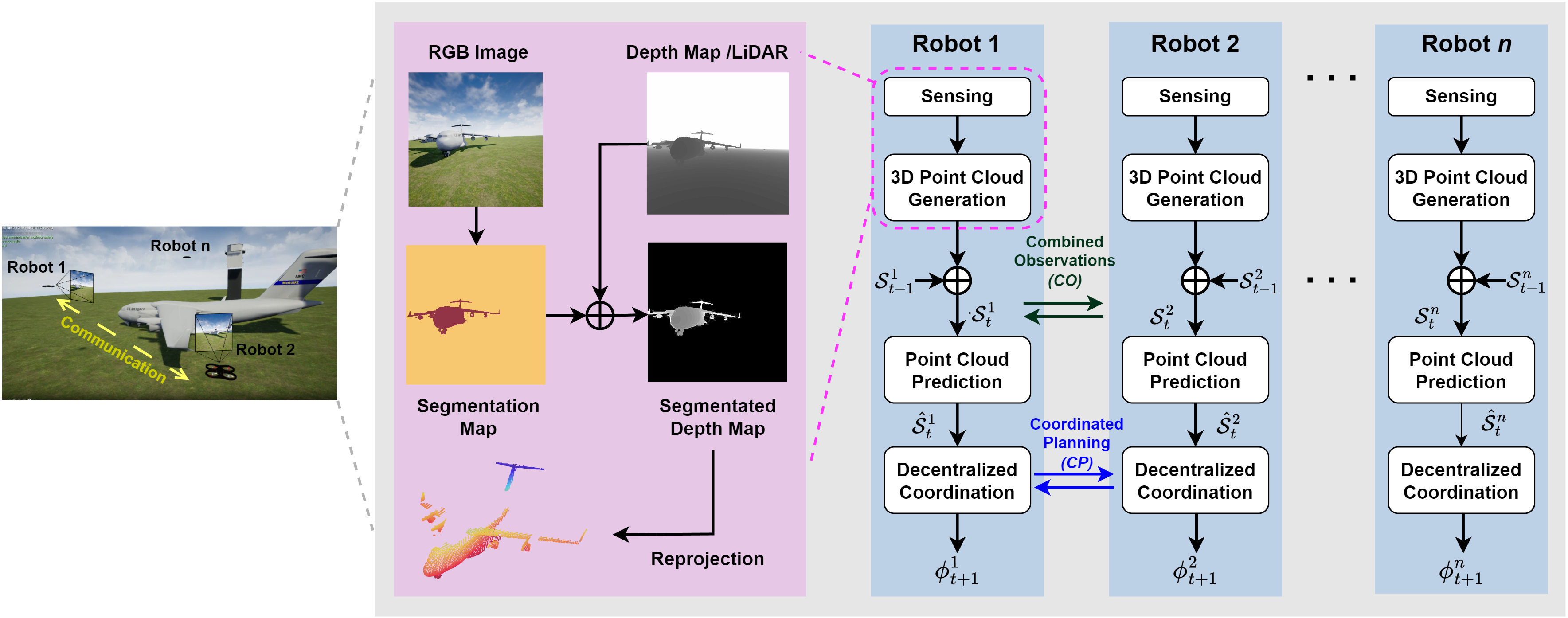}
    \caption{Algorithm Overview: Each robot runs the same algorithm including perception, prediction, and planning steps. The robots that communicate with each other can share observations and coordinate planning, whereas robots in isolation (e.g., Robot n) perform individual greedy planning.}
    \label{fig:algoOverview}
    \vspace{-5mm}
\end{figure*}

\section{Related Work}\label{sec:rel_work}

The use of robots for data acquisition purposes is an extensively studied topic for various domains. Their usage range from infrastructure inspection~\cite{ozaslaninspection} and environment monitoring~\cite{dunbabin2012environmental,sung2019competitive} for real-world application to the real-world digitization for research datasets and simulations~\cite{46965,ramakrishnan2021hm3d,ammirato2017dataset}. When the environment is unknown, active methods such as NBV~\cite{connolly1985determination} are used to construct an object model on the fly by capturing additional observations. A majority of the works on NBV planning use information-theoretic measures~\cite{delmerico2018comparison} for selection to account for uncertainty in observations~\cite{delmerico2018comparison,kuipers1991robot,vasquez2014volumetric}. The widely used frontier and tree-based exploration approaches also utilize uncertainty about the environment for guiding the robot motion~\cite{yamauchi1997frontier, gonzalez2002navigation, adler2014autonomous, bircher2018receding}. Some works devise geometric methods that make inferences about the exact shape of the object of interest and try to align the observations with the inferred model~\cite{tarabanis1995survey, banta2000next, kriegel2013combining}. Prediction-based NBV approaches have emerged as another alternative in recent years, where a neural network takes the robot and/or the environment state as the input and NBV pose or velocity as the output~\cite{johns2016pairwise, mendoza2020supervised, zeng2020pc, dhami2023prednbv}.

Directly extending single-robot NBV approaches to multi-robot systems may result in sub-optimal performance due to significant overlap in observations. This issue led to the development of exploration algorithms specifically for multi-robot systems~\cite{burgard2005coordinated, amanatiadis2013multi, hardouin2020next} with information-theoretic measures for determining NBV. 

Some recent works on multi-robot systems have explored the use of predictions for improvement in task efficiency. Almadhoun et al.~\cite{ almadhoun2021multi} designed a hybrid planner that switches between a classical NBV approach and a learning-based predictor for NBV selection but uses the partial model obtained by robot observations only.
Wu et al.~\cite{wu2019plant} use a point cloud prediction model for plants to use the predicted point cloud as an oracle leading to better results than the traditional approaches. This method uses entropy-based information gain measures for NBV and is designed for plant phenotyping with robotic arms. These methods do not consider the control effort required which is important for UAVs with energy constraints when deployed for observing large objects such as airplanes and ships. Also, these works employ information theoretic NBV approaches. We aim to explore a prediction-based approach for geometric NBV selection.\looseness=-1

In this work, we used point cloud predictions similar to Pred-NBV~\cite{dhami2023prednbv} and built a decentralized multi-robot NBV planner. The prediction on the point cloud makes the pipeline modular and interpretable, allowing for improvements by enhancing individual modules. We select NBV based on information gain, as well as control effort, making our approach more grounded in the real world.

\section{Problem Formulation}\label{sec:prob_form}
We are given a team of \textit{n} robots, each equipped with an RGB-D camera or LiDAR sensor. 
This team is tasked with navigating around a closed object of volume $\mathcal{V} \in \mathbb{R}^3$ and observes its surface as a set of 3D points, $\mathcal{S}$.
At any given time $t$, the set of surface points $S^i_t$ observed by the robot $r_i$ from the view-point $\phi^i_t \in \Phi$ at time $t$ is represented as a voxel-filtered point cloud and the relationship between them is defined as $S^i_t = f(r_i, \phi^i_t)$. Each robot $r_i$ follows a trajectory $\xi_{i}$, which consists of a sequence of viewpoints aimed at maximizing the coverage of $\mathcal{S}$ while minimizing redundancy among observed points. 

The distance traveled by a robot between two poses $\phi_i$ and $\phi_j$ is represented by $d(\phi_i, \phi_j)$. The point cloud observed by the team of robots is the union of the surface points observed by the individual robots over their respective trajectories, i.e., $S_{\Bar{\xi}} =  \bigcup_{i=1}^n \bigcup_{\phi \in \xi_{i}} f(r_i, \phi)$ and $\Bar{\xi}$ represents the set of trajectories for each robot, i.e., $\Bar{\xi} = \{\xi_{1}, \xi_{2},..., \xi_{n}\}$.

The objective is to find a set of feasible trajectories $\Bar{\xi}^* = \{ \xi_{1}^*, \xi_{2}^*, ..., \xi_{n}^*\}$, such that the team observes the whole voxel-filtered surface $\mathcal{S}$, while also minimizing the total distance traveled by the robots on their respective trajectories.
\begin{align}
    \Bar{\xi}^* = \argmin_{\Bar{\xi}} \sum_{i=1}^n \sum_{j=1}^{| {\xi_{i}}-1|} d(\phi_j^i, \phi_{j+1}^i)\\ 
    \textit{such that}~~ \bigcup_{i=1}^n \bigcup_{\phi \in \xi_{i}} f(r_i, \phi) = \mathcal{S} 
\end{align}

Given a finite set of trajectories, if the object model, $\mathcal{S}$, is known, we can find the optimal set of trajectories through an exhaustive search. As the object model is not known apriori in an unknown environment, the optimal solution can not be found beforehand. Thus, each robot needs to determine the NBV based on the partial observations of the team to reconstruct the object's surface. Here we assume that each robot can observe the object at the start of the mission, which can be accomplished by moving the robots till they see the object. While a centralized server can help find an optimal assignment solution, a limited communication range can make a centralized solution infeasible. Thus, we define this problem as a decentralized one; each robot solves this objective but the communicating robots can collaborate and coordinate with their neighbors.

\section{Proposed Approach}\label{sec:approach}
In this section, we present \textit{Multi-Agent Pred-NBV (MAP-NBV)}, a prediction-guided NBV approach for a team of robots. Figure~\ref{fig:algoOverview} shows the overview of our process, which consists of two parts: (1) \textit{3D Model Prediction}, where we combine the observations from the neighboring robots to build a partial model of the object and use PoinTr-C~\cite{dhami2023prednbv}, a 3D point cloud completion network, to predict the full shape of the objects, 
and (2) \textit{Decentralized Coordination} which combines the observations from the communicating robots and solves an NBV objective to maximize information gain and minimize the control effort with sequential greedy assignment. Robots that do not communicate with anyone effectively run an individual greedy algorithm. This approach is detailed in Algorithm~\ref{algo:mapnbv}. Apart from being feasible and scalable, this approach also reduces the computation complexity resulting in fast runtime.

\begin{algorithm}
\caption{MAP-NBV Algorithm (for robot $r_i$)}
\begin{algorithmic}[1]
    \State \textbf{Inputs:} Initial positions $\phi^i_0$; Stopping threshold $\tau$; Information gain threshold $\lambda$
    \State \textbf{Output:} 3D Point Cloud $\mathcal{S}$
    \State \textbf{Initialization}: $novelty$ $\gets 0$; $t \gets 0$; $S^i_{-1} \gets \emptyset$, $\xi_i \gets \{\phi^i_0\}$
    \While{$novelty \le \tau$} \label{algo:line:stopping_condition}
        \State $\mathcal{S}^i_t \gets \bigcup_{k \in Neighbors} f(r_i, \phi^i_t)$ $\bigcup \mathcal{S}^i_{t-1}$  \label{algo:line:combined_observations}
        \State $\hat{\mathcal{S}}^i_t \gets$ getPredictionFromPoinTr-C($\mathcal{S}^i_t$) \label{algo:line:pc_completion}
        \State $\mathcal{C} \gets$ generateCandidatePoses($\hat{\mathcal{S}}^i_t, \mathcal{S}^i_t$) \label{algo:line:cand_pose_generation}

        \State $\mathcal{I}_{seen} \gets \bigcup_{k \in Neighbors; k < i} I (\xi_k)$ \label{algo:line:infogather}

        \State $\hat{\mathcal{I}} \gets \{ \mathcal{I}(\xi_i \cup \phi) - \mathcal{I}_{seen}; \forall  \phi \in \mathcal{C} \}$  \label{algo:line:potential_I_calculation}

        \State $\phi^i_{t+1} \gets \argmin_{\phi \in \mathcal{C}} d(\phi, \phi^i_{t+1})$, s.t. $\frac{\hat{\mathcal{I}}(\phi)}{\max \hat{\mathcal{I}}} \ge \lambda$ \label{algo:line:objective}
        
        \State $\xi_i \gets \xi_i \cup \phi^i_{t+1}$ \label{algo:line:traj_update}
        
        \State Broadcast($\xi_i$) \label{algo:line:broadcast}
        
        \State $novelty \gets \frac{| \mathcal{S}^i_t |}{| \mathcal{S}^i_{t-1}|}$ \label{algo:line:novelty_update}
        \State $t \gets t+1$ \label{algo:line:time_update}
    \EndWhile
    \State $\mathcal{S} \gets \bigcup_{i=1}^n \mathcal{S}^i_{t-1}$  \label{algo:line:novelty_update}\label{algo:line:final_pcd}
\end{algorithmic}
\label{algo:mapnbv}
\end{algorithm}

\subsection{3D Model Prediction (Line~\ref{algo:line:combined_observations}-\ref{algo:line:pc_completion})} 
\VS{To start, we use the RGB images to segment out the object and the depth sensors to generate the point cloud from the current observations for each robot, giving us segmented point clouds. This allows the algorithm to focus on only the target infrastructure as opposed to also including other obstacles. For the robots that can communicate with each other (e.g., Robot 1 and Robot 2 in Fig~\ref{fig:algoOverview}), each segmented point cloud per robot is transformed into a global reference frame and concatenated together into a single point cloud (Line~\ref{algo:line:combined_observations}). This point cloud represents the entire communication subgraph's observations of the target object at the current timestamp.} The point cloud concatenation can be replaced with a registration algorithm~\cite{huang2021comprehensive}, but we use concatenation due to its ease of use. Lastly, this current timestamp's point cloud is then concatenated with previous observations to get an up-to-date observation point cloud.

\VS{In order to get an approximation $\mathcal{\hat{S}}$ of the full model $\mathcal{S}$, we use PoinTr-C~\cite{dhami2023prednbv}. The combined observed point cloud of the object at time $t$, $\mathcal{S}_t$ goes as input to PoinTr-C and it predicts the full object point cloud $\mathcal{\hat{S}}_t$ (Line~\ref{algo:line:pc_completion}). PoinTr-C requires isolating the object point clouds from the scene. This can be realized with the help of distance-based filters and state-of-the-art segmentation networks\cite{kirillov2023segment} without any fine-tuning. An example of an input point cloud and a predicted point cloud, both from individual observations and combined observations, is shown in Figure~\ref{fig:c17predicted}}.

\subsection{Decentralized Coordination (Line~\ref{algo:line:cand_pose_generation}-\ref{algo:line:broadcast})}
\textbf{Next Best View Planning}. We use the predicted point cloud $\mathcal{S}_t$ as an approximation of the ground truth point cloud for NBV planning. For this, we first generate a set of candidate poses around the partially observed object (Line~\ref{algo:line:cand_pose_generation}). From these, we select a set of poses as NBVs for each robot, based on information gain and control effort. For a set of $k$ candidate viewpoints, we define the information gain, $\hat{\mathcal{I}}$, as the expected number of new, unique points the robots will observe after moving to these viewpoints. The control effort is defined as the total distance traversed by the robots to reach the viewpoints.

The number of new points varies with each iteration as robots move to new locations, observing more of the object's surface. While PoinTr-C predicts the point cloud for the whole object, the robots can observe only points on the surface close to it. Hence, before counting the number of new points, we apply hidden point removal~\cite{katz2007direct} to the predicted point cloud. We represent this relationship between the number of points observed and the trajectories traversed till time for a robot $r_i$ as $I(\xi_i)$, where $\xi = \{\phi^i_0, \phi^i_t, ..., \phi^i_t\}$ represents the trajectory for the robot $r_i$ till time $t$ consisting of the set of viewpoints the robot has traversed. To balance the information gain and control effort, we use a hyperparameter $\lambda$. First, we find the estimated information gain for each candidate pose $\hat{\mathcal{I}}$ (Line~\ref{algo:line:potential_I_calculation}) by treating the $\hat{\mathcal{S}}$ as the actual full object model. Each robot selects the candidate pose closest to the current pose where it achieves at least $\lambda \%$ of the maximum possible information gain over all the candidate poses (Line~\ref{algo:line:objective}).\looseness=-1

To find the control effort, we use RRT-Connect~\cite{kuffner2000rrt} to find the path between the robot $r_i$'s current location $\phi^i_t$ to each candidate pose $\phi$ and use its length as $d(\phi, \phi^i_t)$. 
The candidate poses are generated similar to Pred-NBV~\cite{dhami2023prednbv}, i.e., on circles at different heights around the center of the predicted object point cloud (Line~\ref{algo:line:cand_pose_generation}).
One circle is at the same height as the predicted object center with radius $1.5 \times d_{max}$, where $d_{max}$ is the maximum distance of a point from the center of the predicted point cloud. The other two circles are located above and below this circle $0.25 \times \text{z-range}$ away, with a radius of $1.2 \times d_{max}$. The viewpoints are located at steps of $30^\circ$ on each circle.\looseness=-1

\begin{figure}[ht!]
    \centering
    \includegraphics[width=0.90\linewidth]{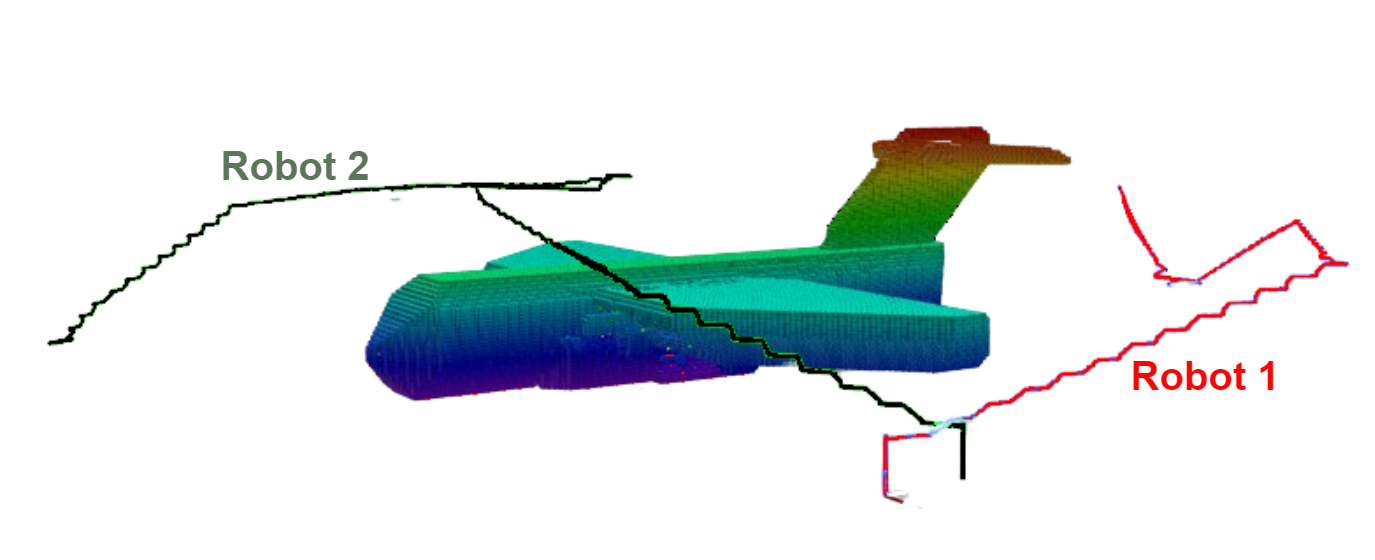}
    \caption{Flight paths of the two robots during C-17 simulation.}
    \label{fig:flightPath}
\end{figure}

\textbf{Sequential Greedy Assignment}. To effectively and efficiently use multiple robots, we coordinate among them for NBV assignment. We devise a decentralized coordination strategy to accommodate the dynamic communication graph effectively. We also leverage the submodular nature of the problem and use a sequential greedy assignment which results in a near-optimal solution~\cite{calinescu2011maximizing}. With each communications subgraph, the robots select their NBVs in the order determined by their IDs (lower to higher). Each robot first considers the trajectories of its neighbors with lower IDs to find the information gain attained by their movement $\mathcal{I}_{seen}$ (Line~\ref{algo:line:infogather}). For the robot with the lowest ID, $\mathcal{I}_{seen}$ is empty. Then we greedily choose the candidate position that would result in the information gain above a threshold $\lambda$ while minimizing the distance traveled. This candidate position for the next iteration 
$\phi_{t+1}^i$ is the NBV for $r_i$. We add this pose to the robot's trajectory (Line~\ref{algo:line:traj_update}) and broadcast this information with the neighbors (Line~\ref{algo:line:broadcast}) to minimize overlaps.

In our experiments, we consider multi-hop communication, thus each robot can coordinate with every robot on its communication subgraph. Some robots may not be within any other robot's communication range (e.g., Robot \textit{n} in Fig~\ref{fig:algoOverview}). For such robots, this strategy effectively turns into a greedy prediction-guided assignment similar to Pred-NBV~\cite{dhami2023prednbv}. At each iteration, we calculate $novelty$, i.e., the ratio of the number of points observed over subsequent iterations (Line~\ref{algo:line:novelty_update}. The robot stops if $novelty$ is above a predefined threshold $\tau$ (we set $\tau = 0.95$ in our experiments).
In the end, all the robots assemble at the same location and combine their observations as the object model $\mathcal{S}$ (Line~\ref{algo:line:final_pcd}).\looseness=-1

\section{Experiments and Evaluation}\label{sec:eval}
We design experiments to answer the two key research questions: (1) can point cloud prediction improve multi-agent object reconstruction? and (2) how does coordination between agents affect the reconstruction? To answer the first question, we compare \textit{MAP-NBV} with a non-predictive frontier-based baseline. For the second question, we compare \textit{MAP-NBV} with a centralized and a non-coordinated variation of \textit{MAP-NBV}. We first describe the experiment setups used to answer these questions and then discuss the results obtained.

\subsection{Setup}
\label{subsec:setup}
\textbf{AirSim}: We use AirSim~\cite{airsim2017fsr} simulator as it allows us to load desired object models and get photo-realistic inputs while also supporting multiple robots. We use Robot Operating System (ROS) Melodic to run the simulations on Ubuntu 18.04. We spawn multiple UAVs close to each other looking towards the object. We equip each UAV with a depth camera and an RGB camera. Each UAV publishes a segmented image using AirSim's built-in segmentation. This segmented image is used along with a depth map to remove the background and isolate the depth map for the object. We then convert these segmented depth images into 3D point clouds.  For collision-free point-to-point planning, we use the MoveIt~\cite{coleman2014reducing} package implementing the work done by Köse~\cite{tahsinko86:online}. 


\begin{figure}[htp]
\begin{minipage}[b]{0.43\linewidth}
    \centering
    \begin{subfigure}[b]{\linewidth}
        \centering
        \includegraphics[width=\textwidth]{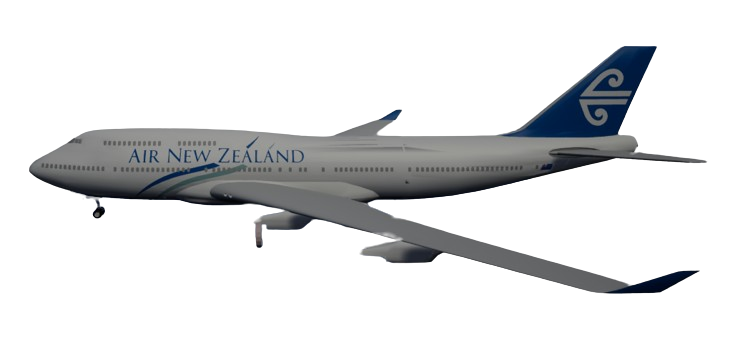}
        \caption{Airplane}
        \label{fig:airplane}
    \end{subfigure}
    

    \begin{subfigure}[b]{\linewidth}
        \centering
        \includegraphics[width=\linewidth]{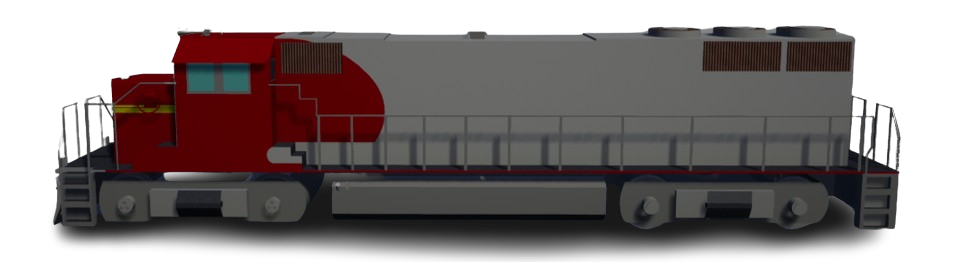}
        \caption{Train}
        \label{fig:train}
    \end{subfigure}


    \begin{subfigure}[b]{\linewidth}
        \centering
        \includegraphics[width=\linewidth]{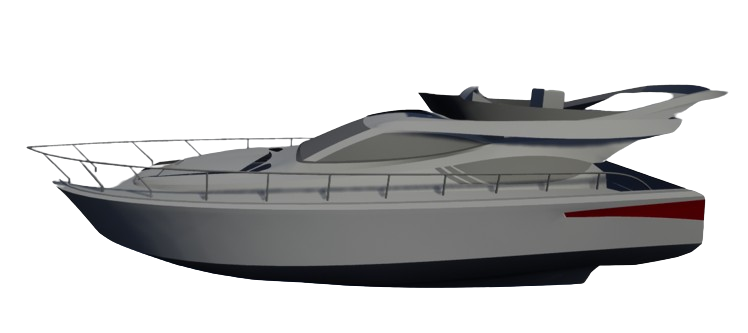}
        \caption{Boat}
        \label{fig:boat}
    \end{subfigure}
\end{minipage}
\begin{minipage}[b]{0.56\linewidth}
    \centering
    \begin{subfigure}[b]{0.41\linewidth}
        \centering
        \includegraphics[width=0.80\textwidth]{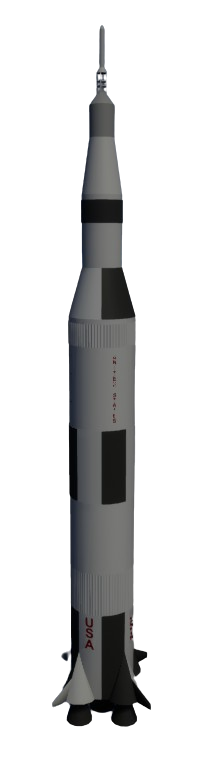}
        \caption{Rocket}
        \label{fig:rocket}
    \end{subfigure}
    \begin{subfigure}[b]{0.55\linewidth}
        \centering
        \includegraphics[width=0.90\textwidth]{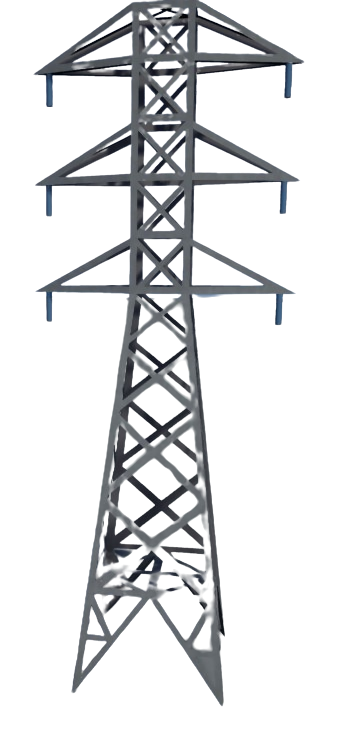}
        \caption{Tower}
        \label{fig:tower}
    \end{subfigure}
\end{minipage}

\caption{Examples of the 5 simulation model classes.}
\label{fig:simModels}
\end{figure}

\textbf{ShapeNet}: The AirSim setup helps us evaluate the coverage in a realistic setting, but it lacks a ground truth point cloud to compare the reconstruction quality. Hence, we run simulations with the ShapeNet~\cite{shapenet2015} dataset to obtain a ground truth and compare the reconstruction quality across different methods. We simulate the UAVs as free-floating cameras which can teleport to any location. These cameras do not have a roll and pitch and can only change their yaw. We use the Open3D~\cite{Zhou2018} library to load an object mesh (normalized) and select a random starting position around it. The UAVs are placed close to each other at this location. We render the depth image from the object mesh and reproject it to the 3D point cloud. Unlike the previous setup, we do not need segmentation as there is no background in this setup. 
To evaluate the reconstruction quality, we convert the mesh to a point cloud by uniformly sampling points on it and use it as the ground truth to quantify the reconstruction quality with Directional Chamfer Distance with $\ell_2$-norm (CD-$\ell_2$) (from ground truth point cloud to the observed point cloud). We use Euclidean distance as the distance metric $d(.,.)$ in this setup.\looseness=-1

\subsection{Qualitative Example}
We evaluate \textit{MAP-NBV} on the same 20 objects that were used in Pred-NBV to allow a direct comparison. 
The 20 objects consist of five different ShapeNet classes:  airplane, rocket, tower, train, and watercraft. Examples of each class are shown in Figure~\ref{fig:simModels}. These classes represent diverse shapes and infrastructures that are regularly inspected. Figure~\ref{fig:flightPath} shows the path followed by two UAVs as given by \textit{MAP-NBV} in the C-17 airplane simulation. This environment includes other obstacles that are not of interest but still need to be accounted for in collision-free path planning. \textit{MAP-NBV} finds a collision-free path for both UAVs while targeting the maximum coverage of the C-17 airplane.

\subsection{Can Point Cloud Prediction Improve Reconstruction?}\label{sec:sim:MAbaseline}
We compared the performance of \textit{MAP-NBV} with a modified baseline NBV method~\cite{aleotti2014global} designed for multi-agent use on 20 objects, as listed in Table~\ref{tab:airsim_multiagent_results}. The baseline method employs frontiers to select the next-best views. Frontiers are points located at the edge of the observed space near unknown areas. We utilized the same modifications described in Pred-NBV~\cite{dhami2023prednbv}. Specifically, we used our segmented point cloud to choose frontiers near the target object. To ensure that the UAVs always face the target object, the orientation of all poses selected by the baseline aligns with the center of the observed target object point clouds.

We further adapted this baseline method to function in a multi-agent setting. The pose for the first UAV is selected in the exact same manner as in the single-agent baseline. For each subsequent UAV, the remaining best pose is chosen, as long as it does not fall within a certain distance threshold compared to the previously selected poses in the current iteration of the algorithm.

Both \textit{MAP-NBV} and the baseline algorithm employ the same stopping criteria. The algorithm terminates if the total points observed in the previous step exceed 95\% of the total points observed in the current step. We run both the algorithms in the AirSim setup. Our evaluation, presented in Table~\ref{tab:airsim_multiagent_results}, demonstrates that \textit{MAP-NBV} observes, on average, 19.41\% more points than the multi-agent baseline for object reconstruction across all 20 objects from the five different model classes. In our simulations, we utilized 2 UAVs for both algorithms.\looseness=-1

Furthermore, the \textit{MAP-NBV} algorithm can be readily extended to accommodate more than just 2 robots. By incorporating additional UAVs, the algorithm can effectively leverage the collaborative efforts of a larger multi-agent system to improve object reconstruction performance and exploration efficiency. However, in our current evaluation, we utilized 2 UAVs for both algorithms due to limited computational resources. The simulations were computationally intensive, and our computer experienced significant slowdowns with just 2 robots in the simulation. Despite this limitation, the promising results obtained with 2 UAVs suggest that scaling up the algorithm to include more robots has the potential to yield even more significant performance improvements.

\begin{figure}[ht!]
    \vspace{-3mm}
    \centering
    \includegraphics[width=0.93\linewidth]{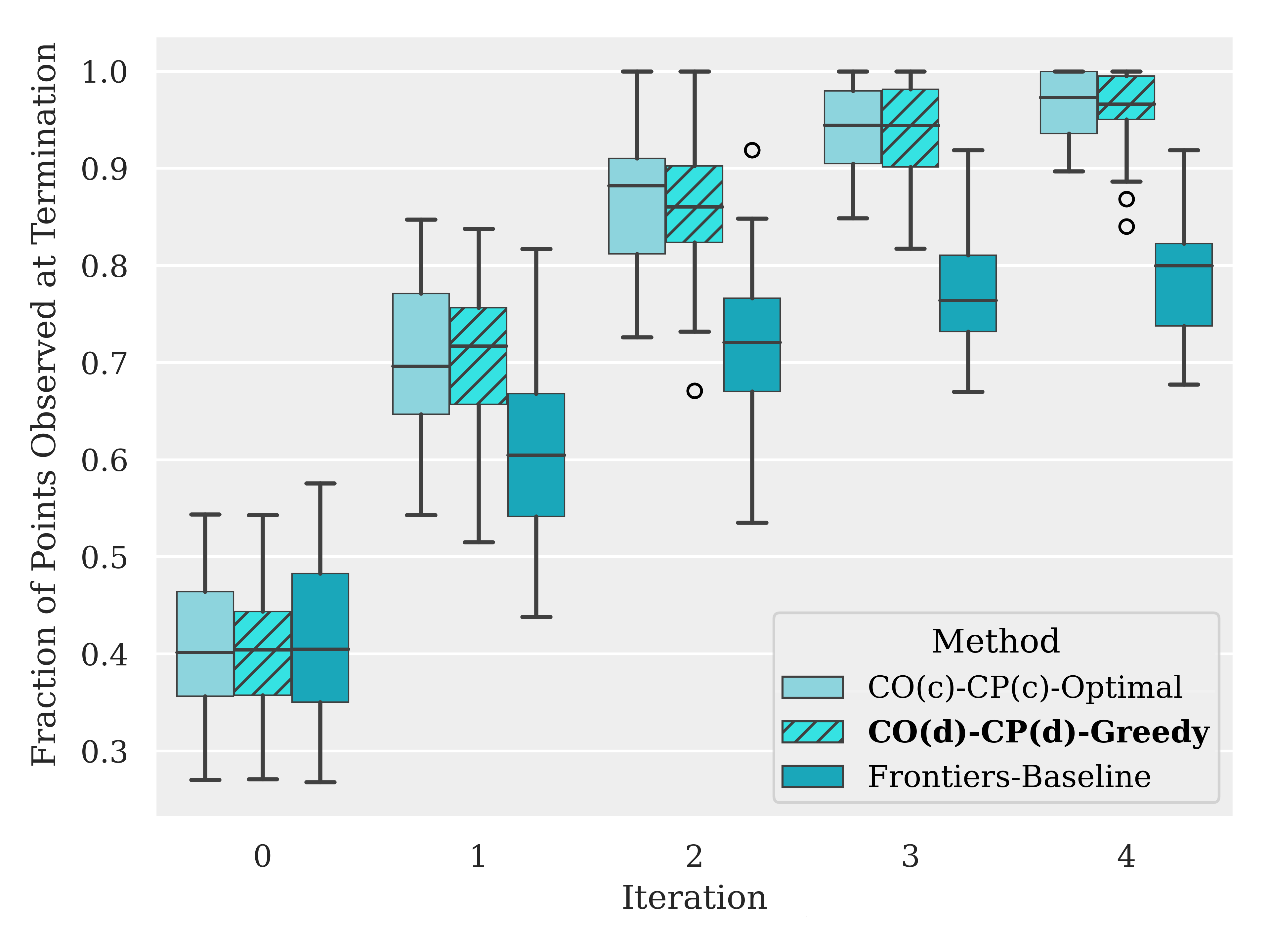}
    \caption{MAP-NBV (\texttt{CO(d)-CP(d)-Greedy}) performs comparably to the optimal solution (\texttt{CO(c)-CP(c)-Optimal}; Section~\ref{sec:COCPdesc}),  and much better than the frontiers-baseline in AirSim experiments.\looseness=-5 }
    \label{fig:plane_res}
\end{figure}

Additionally, Figure~\ref{fig:plane_res} illustrates that \textbf{\textit{MAP-NBV} observes more points per step than the multi-agent baseline}.

\begin{table}[ht!]
\vspace{1.0mm}
    \centering

    \caption{\textit{MAP-NBV} results in a better coverage compared to the multi-agent baseline NBV method~\cite{aleotti2014global} for all models in AirSim upon algorithm termination.}
    
    \begin{tabular}{p{0.9cm}p{1.3cm}rrr}
    \toprule
    \multirow{2}{*}{Class} & \multirow{2}{*}{Model} & \multicolumn{2}{c}{Points Seen} & \multirow{2}{*}{Improvement} \\
    \cmidrule(lr){3-4}
    & & \textit{MAP-NBV} & MA Baseline & \\
        \midrule
        \multirow{5}{*}{Airplane}  
            & 747 & \textbf{16140} & 13305 & 19.26\% \\
            & A340 & \textbf{10210} & 8156 &  22.37\% \\
            & C-17 & \textbf{13278} & 10150 & 26.70\% \\
            & C-130 & \textbf{6573} & 5961 & 9.77\% \\
            & Fokker 100 & \textbf{14986} & 13158& 12.99\%\\
        \midrule
        \multirow{5}{*}{Rocket} 
            & Atlas & \textbf{2085} & 1747 & 17.64\% \\
            & Maverick & \textbf{3625} & 2693 & 29.50\% \\
            & Saturn V & \textbf{1041} & 877 & 17.10\% \\
            & Sparrow & \textbf{1893} & 1664 & 12.88\% \\
            & V2 & \textbf{1255} & 919 & 30.91\% \\
        \midrule
        \multirow{5}{*}{Tower} 
            & Big Ben & \textbf{4294} & 3493 & 20.57\% \\
            & Church & \textbf{7884} & 6890 & 13.46\% \\
            & Clock & \textbf{3163} & 2382 & 28.17\% \\
            & Pylon & \textbf{2986} & 2870 & 3.96\% \\
            & Silo & \textbf{5810} & 4296 & 29.96\% \\
        \midrule
        \multirow{2}{*}{Train}
            & Diesel & \textbf{4013} & 3233 & 21.53\% \\
            & Mountain & \textbf{5067} & 4215 & 18.36\% \\
        \midrule
        \multirow{3}{*}{Watercraft} 
            & Cruise & \textbf{5021} & 3685 & 30.69\% \\
            & Patrol & \textbf{4078} & 3683 & 10.18\% \\
            & Yacht & \textbf{11678} & 10341 & 12.14\% \\
        \bottomrule
    \end{tabular}
    
    \label{tab:airsim_multiagent_results}
\end{table}

\subsection{How does coordination affect reconstruction?}\label{subsec:ablation}
\VS{For studying the effect of coordination we randomly select 25 objects, 5 from each of the ShapeNet classes mentioned in Section~\ref{sec:sim:MAbaseline} for the ShapeNet setup (Section~\ref{subsec:setup}).}

\label{sec:COCPdesc}
In a multi-agent setting, the communicating robots can combine their observation (\textbf{CO}) or they may choose to rely on their observations (\textbf{IO}) for point cloud predictions. After the prediction, they may choose to collaborate for planning (\textbf{CP}) or make individual decisions without relying on others (\textbf{IP}). To highlight the effect of these coordination strategies on reconstruction quality, for a team of $n$ robots and $k$ candidate viewpoints, we compare MAP-NBV with two contrasting coordination approaches: 
\begin{itemize}
    \item \textbf{\texttt{CO(c)-CP(c)-Optimal}}: This approach relies on combined observation and coordinated planning in a centralized manner. Here, a central server aggregates the observations, performs prediction, and finds the NBVs for each robot. For selecting NBVs, the algorithm evaluates all robot-candidate pose assignments and chooses the optimal setting, i.e.,  which results in maximum joint information gain. Among all possible permutations of robots that result in the optimal setting, we select the one that minimizes the maximum displacement for any robot. This process has a runtime complexity of $\mathcal{O}(k^{n})$.
    \item \textbf{\texttt{IO-IP}}: In this approach, each robot operates individually and does not share observations or coordinate with others regardless of the communication range. This is effectively a naive extension of Pred-NBV to a multi-agent setup. The runtime complexity here is $\mathcal{O}({k})$.
\end{itemize}

Following the notations above, \textbf{\uline{MAP-NBV is \texttt{CO(d)-CP(d)-Greedy}}}, i.e., combined observations (decentralized) and coordinated planning (decentralized) with greedy assignment. \textit{Decentralized} indicates that the observations and planning are shared among only those robots that form a communication subgraph. \textit{MAP-NBV} thus has a runtime complexity of $\mathcal{O}(n \cdot k)$ in the worst case only, i.e., when all the robots are connected. We study these algorithms for teams of 2, 4 and 6 robots. 

\begin{figure}[ht!]
    \vspace{-2mm}
    \centering
    \begin{subfigure}[b]{\columnwidth}%
        \includegraphics[width = 0.95\textwidth]{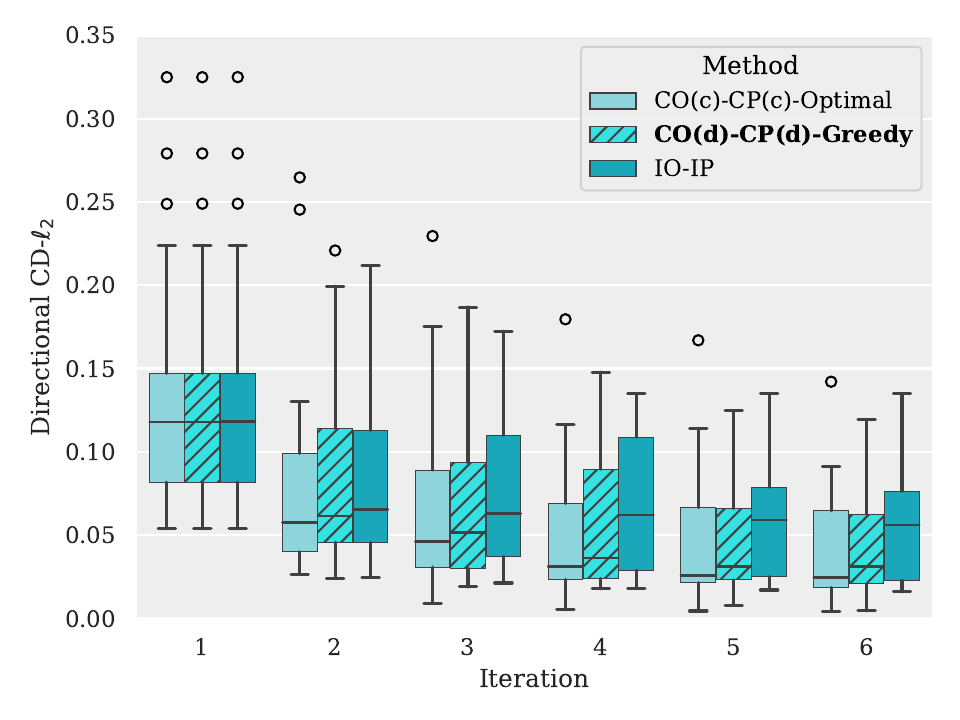}%
        \vspace{-3mm}
        \subcaption{2 Robots}
    \end{subfigure}%
    \hfill%
    \begin{subfigure}[b]{\columnwidth}%
        \includegraphics[width = 0.95\textwidth]{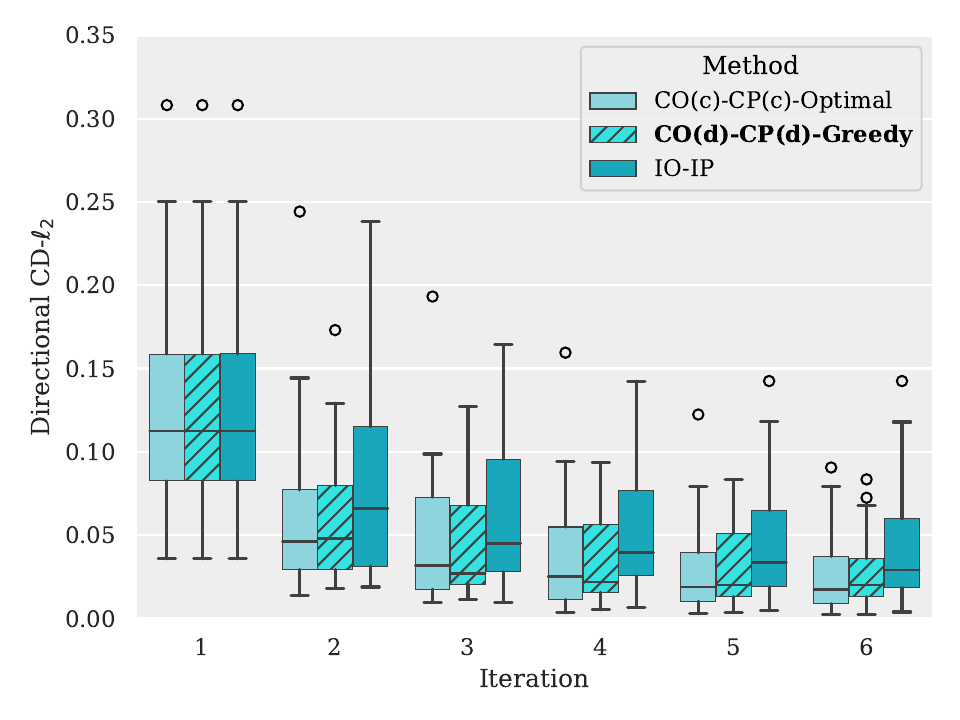}%
        \vspace{-3mm}
        \subcaption{4 Robots}
        
    \end{subfigure}%
    \hfill%
    \begin{subfigure}[b]{\columnwidth}%
        \includegraphics[width = 0.95\textwidth]{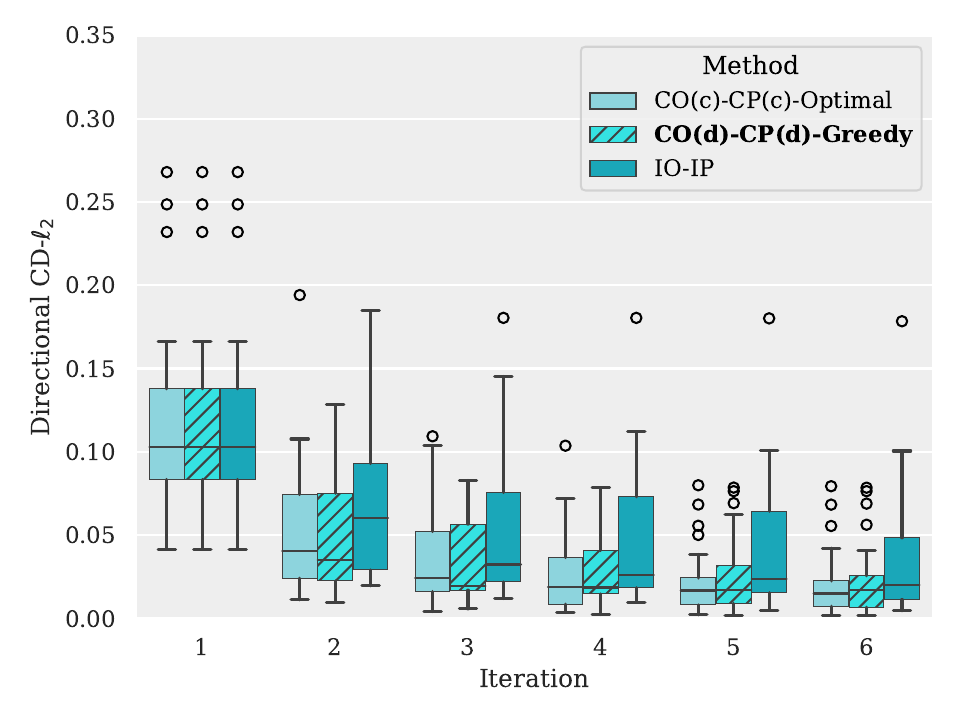}%
        \vspace{-3mm}
        \subcaption{6 Robots}
    \end{subfigure}%
    \caption{Directional CD-$\ell_2$ for teams of 2, 4, and 6 robots on ShapeNet models~\cite{shapenet2015} with different coordination strategies.}
    \label{fig:ablation}
    \vspace{-1mm}
\end{figure}

At each iteration, we combine the observations from all robots and compute Direction CD-$\ell_2$ over it. This thus represents the reconstruction quality if the mission terminated at that iteration. We found that 6 iterations were enough for all robots to reach the stopping criteria in each setting. Our findings are shown in Figure~\ref{fig:ablation}.

We observed that coordination and information sharing play a crucial role in improving the reconstruction quality. \texttt{CO(c)-CP(c)-Optimal} exhibits the best reconstruction over time, as expected, owing to shared observations which lead to better estimation of the partial point cloud, and coordinated planning, which minimizes overlap in information gain. In early iterations, the limited observations may lead to an imprecise prediction, which may result in inefficient NBV assignments. Still, over time the observation coverage increases leading to better predictions and performance over the other algorithms. We observed similar trends for the AirSim setup as well as shown in Figure~\ref{fig:plane_res}.

However, \texttt{CO(c)-CP(c)-Optimal} requires large computation time and does not scale well in comparison with \textit{MAP-NBV} \texttt{(CO(d)-CP(d)-Greedy)}. On a Ubuntu 20.04 system with 32-core, 2.10Ghz Xeon Silver-4208 CPU and Nvidia GeForce RTX 2080Ti GPU, \textit{MAP-NBV} was 2x faster for 4 robots and 10x faster for 6 robots compared to \texttt{CO(c)-CP(c)-Optimal}. Even with faster execution, \textit{MAP-NBV} performs comparably to \texttt{CO(c)-CP(c)-Optimal}, making it more attractive than the former. Additionally, the decentralized nature makes \textit{MAP-NBV} a more feasible algorithm than \texttt{CO(c)-CP(c)-Optimal} which requires centralization. \texttt{IO-IP}, where the robots do not share observations or coordinate plans, exhibits the worst improvement over time, highlighting that \textbf{the coordination plays a crucial role in multi-agent object reconstruction}. In fact, \textit{MAP-NBV} achieves a relative improvement of \textbf{17-22\%} in directional CD-$\ell_2$ over \texttt{IO-IP} after termination.

Interestingly, we observe that as the number of robots increases, the gap between these algorithms decreases. This is expected as more robots lead to more coverage at any time. Since the robots start the mission at close distances in our experiments, we found that \texttt{IO-IP} still exhibits worse improvement as the robots suffer significant overlaps in observations. Thus coordination still provides benefits and faster runtime of \textit{MAP-NBV} makes it a more suitable choice than the centralized alternative, even if centralization is feasible. 
We also performed experiments with a centralized, greedy selection variant of \textit{MAP-NBV} and found it performs only marginally better than \textit{MAP-NBV}'s performance. We share these findings on our \href{https://raaslab.org/projects/MAPNBV/}{project webpage} due to lack of space.\looseness=-1

\subsection{Qualitative Real-World Experiment}
\begin{figure*}[ht!]
    \centering
    \begin{subfigure}[c]{.24\textwidth}%
        \includegraphics[width = \textwidth]{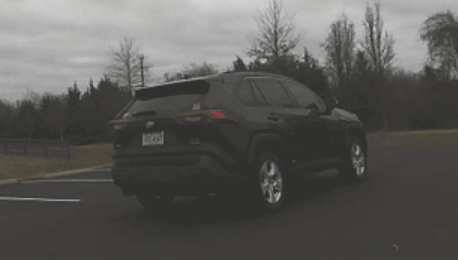}%
        \caption{}
    \end{subfigure}%
    \hfill
    \begin{subfigure}[c]{.24\textwidth}%
        \includegraphics[width = \textwidth]{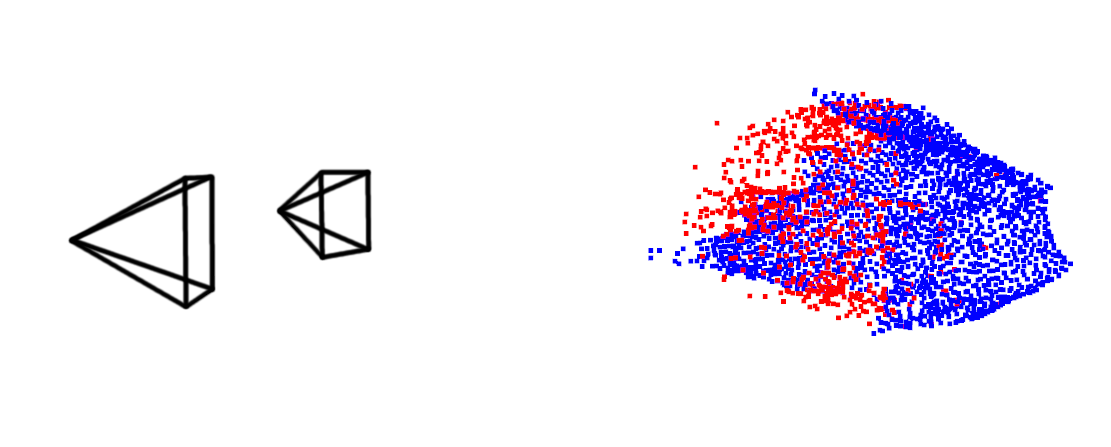}%
        \caption{}
    \end{subfigure}%
    \hfill
    \begin{subfigure}[c]{.24\textwidth}%
        \includegraphics[width = \textwidth]{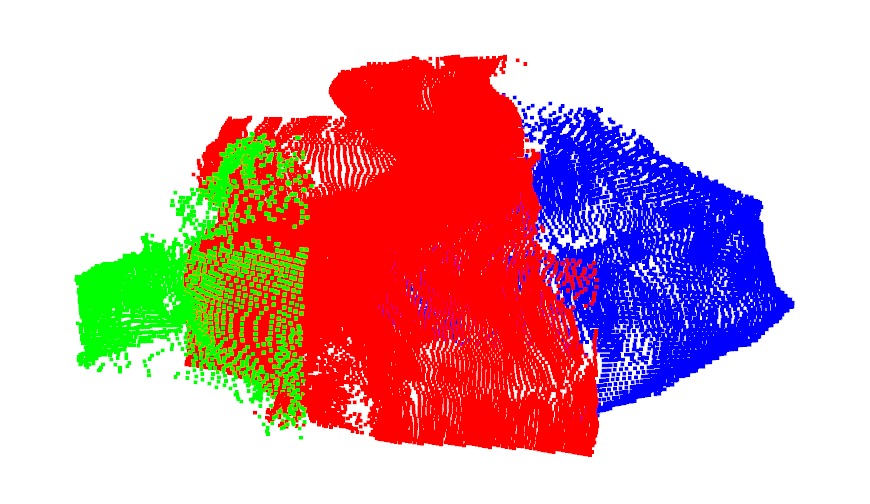}%
        \caption{}
    \end{subfigure}%
    \hfill
    \begin{subfigure}[c]{.24\textwidth}%
        \includegraphics[width = \textwidth]{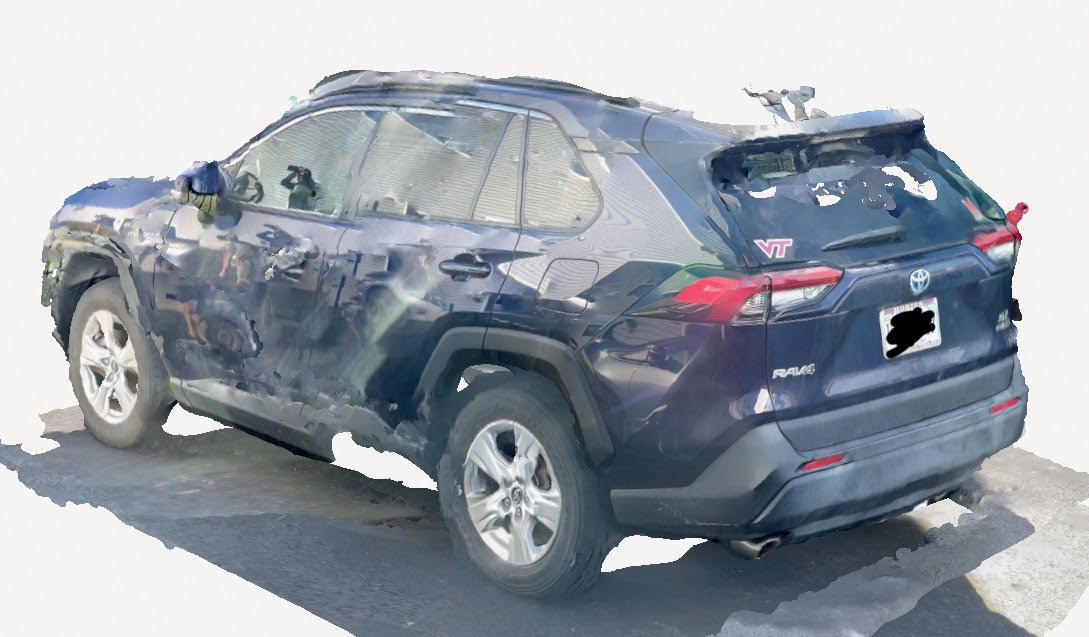}%
        \caption{}
    \end{subfigure}
    \caption{Real-World MAP-NBV experiment. (a) RGB Image. (b) \textcolor{blue}{Observations}, \textcolor{red}{Predictions}, and MAP-NBV poses. (c) \textcolor{blue}{Initial}, \textcolor{red}{Drone 1}, and \textcolor{green}{Drone 2} points after MAP-NBV iteration. (d) Reconstruction.}
    \label{fig:zed_experiment}
    \vspace{-5mm}
\end{figure*}

We also conducted a real-world experiment to evaluate the feasibility of implementing MAP-NBV on hardware. A single iteration of the MAP-NBV pipeline for 2 robots was run using a ZED camera. Depth images and point clouds were captured from the ZED of a Toyota RAV4 car. The car was extracted from the point clouds and used as input for the MAP-NBV pipeline. Then, depth images and point clouds were captured from the poses MAP-NBV outputted. Due to the lower quality of the ZED camera, an iPhone 12 Pro Max with built-in LiDAR was used to capture data to create a high-resolution reconstruction shown in Figure~\ref{fig:zed_experiment}d.

\section{Conclusions}\label{sec:con}
We present a multi-agent, decentralized, prediction-guided NBV planning approach for active 3D reconstruction. This method can be helpful in a variety of applications including civil infrastructure inspection. We show that our method can faithfully reconstruct the object point clouds efficiently compared to non-predictive multi-agent methods and other prediction-guided approaches. Our NBV planning objective considers both information gain and control effort, making it more suitable for real-world deployment given the flight time limit imposed on UAVs by their battery capacity. 

We are currently working on a bandwidth-aware extension of \textit{MAP-NBV} and the preliminary studies show encouraging results (shared on our \href{https://raaslab.org/projects/MAPNBV/}{webpage}). In this work, we focus solely on geometric measures for information gain. Many existing works on NBV have developed sophisticated information theoretic measures. We will explore combining both types of measures in our future work.

{\small
\bibliographystyle{IEEEtran}
\bibliography{main}
}

\end{document}